\documentclass[journal]{IEEEtran}

\usepackage[cmex10]{amsmath}

\usepackage{amssymb}
\usepackage{graphicx}
\usepackage{multirow}
\usepackage{url}
\usepackage{booktabs}
\usepackage{algorithmic}
\DeclareMathOperator*{\argmax}{arg\,max}

\begin{document}

\title{Remote sensing image classification exploiting
  multiple kernel learning}

\author{Claudio~Cusano, Paolo~Napoletano, Raimondo~Schettini
\thanks{C. Cusano is with Dept. of Electrical, Computer and Biomedical Eng., University of Pavia, Italy. P.~Napoletano and R. Schettini are with Dept. of Informatics, Systems and Comm.,
  University of Milan-Bicocca, Milano, Italy}
}

\maketitle


\begin{abstract}
  We propose a strategy for land use classification which exploits
  Multiple Kernel Learning (MKL) to automatically determine a suitable
  combination of a set of features without requiring any heuristic
  knowledge about the classification task.  We present a novel
  procedure that allows MKL to achieve good performance in the case of
  small training sets.  Experimental results on publicly available
  datasets demonstrate the feasibility of the proposed approach.
\end{abstract}

\begin{IEEEkeywords}
Remote sensing image classification, multiple kernel learning (MKL)
\end{IEEEkeywords}

\section{Introduction}

\IEEEPARstart{T}{he} automatic classification of land use is of great
importance for several applications in agriculture, forestry, weather
forecasting, and urban planning.  Land use classification consists in
assigning semantic labels (such as urban, agricultural, residential
etc.) to aerial or satellite images.  The problem is closely related
to that of land cover classification and differs from it mainly in the
set of labels considered (forest, desert, open water etc.).  In the
past these problems have been often addressed by exploiting spectral
analysis techniques to independently assign the labels to each pixel
in the image~\cite{campbell2002introduction}.  Recently, several
researchers experimented with image-based techniques consisting,
instead, in extracting image features and in classifying them
according to models obtained by supervised learning.  Multiple
features should be considered because the elements in the scene may
appear at different scales and orientations and, due to variable
weather and time of the day, also under different lighting
conditions.

Sheng \emph{et al.} designed a two-stage classifier that combines a
set of complementary features~\cite{sheng2012highresolution}.  In the
first stage Support Vector Machines (SVM) are used to generate
separate probability images using color histograms, local ternary
pattern histogram Fourier (LTP-HF) features, and some features derived
from the Scale Invariant Feature Transform (SIFT).  To obtain the
final result, these probability images are fused by the second stage
of the classifier.
Risojevi{\'c} \emph{et al.} applied non-linear SVMs with a kernel
function especially designed to be used with Gabor and Gist
descriptors~\cite{risojevic2011gabor}.
In a subsequent work, Risojevi\'c and Babi\'c considered two features:
the Enhanced Gabor Texture Descriptor (a global feature based on
cross-correlations between subbands) and a local descriptor based on
the SIFT.  They identified the
classes of images best suited for the two descriptors, and used this
information to design a hierarchical approach for the final
fusion~\cite{risojevic2013fusion}.
Shao~\emph{et al.}~\cite{shao2013hierarchical} employed both
non-linear SVMs and L1-regularized logistic regression in different
classification stages to combine SIFT descriptors, shape features for image
indexing, texture feature based on Local
Binary Patterns (LBP) and a bag-of-colors signature.

In this paper, we propose a strategy for land use classification which
exploits Multiple Kernel Learning (MKL) to automatically combine a set
of features without requiring any heuristic knowledge about the
classification task.  One of the drawbacks of MKL is that it requires
a large training set to select the features and to train the
classifier simultaneously.  In order to apply MKL to small training
sets as well, we introduce a novel automatic procedure that produces
candidate subsets of the available features before solving the
optimization problem defined by the MKL framework.  The proposed
strategy exploits both features commonly used in scene classification
as well as new features specially designed by the authors to cope with
the land use classification problem. We evaluated our proposal on two
data sets: the Land use data set of aerial images~\cite{yang2010bag},
and a data set of satellite images obtained from the Google Earth
service~\cite{dai2011satellite}. We compared our framework with others
from the state of the art.  The results show that the proposed
framework performs better than other methods when the training set is
small.

\section{Proposed classification scheme}

Multiple Kernel Learning (MKL) is a powerful machine learning tool
which allows, in the framework of Support Vector Machines (SVM), to
automatically obtain a suitable combinations of several kernels over
several features (therefore selecting the features as
well)~\cite{lanckriet2004statistical,sonnenburg2006large,tuia2010learning,subrahmanya2010sparse,gu2012representative}.  
We evaluated the MKL framework as proposed by Sonnenburg~\emph{et
  al.}~\cite{kloft2011lp} in the case of small size of the training
set (results are detailed in the experimental section). We observed
that dense mixtures of kernels can fit real data better than sparse
mixtures but also that both sparse and non sparse solutions do not
outperform other trivial baselines solutions such as those proposed by
Risojevi\'c and \emph{et al.}~\cite{risojevic2013fusion}.

To improve the results in the case of small training sets, we
introduce a novel heuristic procedure that automatically selects
candidate subsets of the available combinations of features and
kernels before solving the MKL optimization problem.

\subsection{Multiple Kernel Learning}
Support Vector Machines exploit the `kernel trick' to build non-linear
binary classifiers.  Given a feature vector $\mathbf{x}$, the
predicted class $y$ (either $-1$ or $+1$) depends on a kernel $k$:
\begin{equation}
  \label{eq:svm}
  y = \mathrm{sign}(b  +
  \sum_{i=1}^N \alpha_i y_i k(\mathbf{x}, \mathbf{x}_i)),
\end{equation}
where $(\mathbf{x}_i, y_i)$ are the features/label pairs forming the
training set.  The parameters $b$ and $\alpha_i$ are determined during
the training procedure, which consists in solving the following
quadratic optimization problem:
\begin{equation}
  \label{eq:svmtrain}
  \min_{\mathbf{\alpha}} \frac{1}{2}
  \sum_{i=1}^N  \sum_{j=1}^N \alpha_i \alpha_j y_i y_j
  k(\mathbf{x}_i, \mathbf{x}_j) - \sum_{i=1}^N \alpha_i,
\end{equation}
under the constraints $\sum_i \alpha_i y_i = 0$ and $0 \leq \alpha_i
\leq C$ (for a set value of the penalization parameter $C$).  Beside
the linear kernel ($k(\mathbf{x}_1, \mathbf{x}_2) = \mathbf{x}_1^T
\cdot \mathbf{x}_2$), other popular kernels are the Gaussian RBF
($k(\mathbf{x}_1, \mathbf{x}_2) = \exp( -\gamma \| \mathbf{x}_1 -
\mathbf{x}_2 \|^2)$) and the $\chi^2$ kernel ($k(\mathbf{x}_1,\mathbf{x}_2) = \exp( -\gamma \sum_j \frac{(x_{1j} -  x_{2j})^2}{x_{1j}+ x_{2j}})$) both depending on an additional parameter $\gamma$.

The choice of the kernel function is very important since it
implicitly defines the metric properties of the feature space.
Multiple Kernel Learning (MKL) is an extension of Support Vector
Machines (SVMs) that combines several kernels.  It represents an
appealing strategy
when dealing with multiple feature vectors $\mathbf{x} = \langle
\mathbf{x}^{(1)}, \mathbf{x}^{(2)}, \dots, \mathbf{x}^{(F)} \rangle$
and with multiple kernel functions $k_1^{(f)}, k_2^{(f)}, \dots,
k_M^{(f)}$ for each feature.  In MKL, the kernel function used in
(\ref{eq:svm}) is a linear combination of these $F \times M$ kernels
\begin{equation}
  k_{comb}(\mathbf{x}, \mathbf{x}_i) = \sum_{f=1}^{F} \sum_{j=1}^M \beta^{(f)}_j k_j^{(f)}(\mathbf{x}^{(f)}, {\bf x}_i^{(f)}).
\label{eq:MKL_1}
\end{equation}
The weights $\beta^{(f)}_j \geq 0$ are considered as additional
parameters to be found when solving the optimization problem
(\ref{eq:svmtrain}) with the inclusion of the new constraint
$\sum_f \sum_j \beta^{(f)}_j = 1$.

The first approaches to MKL aimed at finding sparse mixtures of
kernels with the introduction of $l_{1}$ regularization for the
weights of the combination.  For instance, Lanckriet \emph{et al.}
induced the sparsity by imposing an upper bound to the trace of
$k$~\cite{Lanckriet2004}.
The resulting optimization was computationally too expensive for large
scale problems~\cite{kloft2009efficient}. Moreover, the use of sparse
combinations of kernels rarely demonstrated to outperform trivial
baselines in practical applications.

Sonnenburg \emph{et al.} introduced an efficient strategy for solving
$l_{p}$-regularized MKL with arbitrary norms, $p \geq
1$~\cite{kloft2011lp,kloft2009efficient}.  The algorithm they proposed
allows to estimate optimal weights and SVM parameters simultaneously
by iterating training steps of a standard SVM~\cite{kloft2011lp}.
This strategy allows both sparse ($p=1$) and non-sparse ($p>1$)
solutions (for the non-sparse case the constraint on the weights must
be changed to $\sum_f \sum_j \left (\beta_j^{(f)} \right)^p  = 1$).

In the case of small training sets, plain SVMs may outperform
MKL. This fact mostly depends on the increased number of parameters
that have to be set during training, and is particularly evident when
there are many features and kernels.

\subsection{Heuristic MKL}
We introduce, here, a novel heuristic approach to MKL that finds
sparse mixtures of kernels without imposing the sparsity condition
($p=1$).  Using a small number of kernels is very important in the
case of small training sets because it reduces the number of
parameters to be learned and, consequently, limits the risk of
overfitting the training data.  In fact, using all the available
kernels is usually worse than using a small selection of good kernels.
Sparsity could be enforced by constraining to be zero a subset of the
coefficients of the kernels before solving the $l_2$ regularized
optimization problem.  The optimal solution could be found by
considering all the $2^{F \times M}$ possible subsets.  However, this
approach would easily result intractable even for relatively small
numbers of feature vectors ($F$) and kernels ($M$).  A tractable
greedy solution would consist in selecting one kernel at a time on the
basis of its individual merit. This approach, instead, would fail to
capture most of the interactions among the kernels.

Our heuristic strategy deals with the limitations of the greedy
algorithm, without resulting intractable as the exhaustive search.
Briefly, It consists in the automatic selection of a small number of
kernels (one for each feature vector) and in an iterative augmentation
of such an initial selection.  New kernels are not included only
because of their performance in isolation, but they are chosen by taking into account how much
they complement those that have been already selected.  Since
complementarity is more easily found across different features, at
each iteration the algorithm considers for the inclusion at most one
kernel for each feature vector.  More in detail, the procedure is
composed of four major steps. For each step the goodness of one or
more sets of kernels is evaluated by training the MKL with $p=2$ and
by a five-fold cross validation on the training set:
\begin{enumerate}
\item for each of the $F$ feature vectors and for each of the $M$
  kernels, a classifier is trained and evaluated; the best kernel for
  each feature is then included in the initial selection
  $\mathcal{S}$;
\item the inclusion of each non-selected kernel is individually
  evaluated after its temporary addition to $\mathcal{S}$;
\item for each feature vector, the kernel corresponding to the largest
  improvement in step 2 is taken; these kernels are used to form a set
  $\mathcal{C}$ of candidate kernels.  Features whose kernel does not
  improve the accuracy will remain without a candidate;
\item for each subsets of $\mathcal{C}$, its union with $\mathcal{S}$
  is evaluated; the subset $\mathcal{B}^*$ corresponding to the
  largest improvement is permanently added to $\mathcal{S}$.
\end{enumerate}
The steps 2--4 are repeated until the set of candidates $\mathcal{C}$
found in step 3 is empty (this would eventually happen since each step
adds at least one kernel until no kernel improves the
accuracy, or until all the kernels have been selected).
\begin{figure}[tb]
\centering
  \footnotesize
  \begin{algorithmic}
    \newcommand\eval[1]{\ensuremath{\text{{\bf evaluate}}}(#1)}
    \newcommand\A{\leftarrow}

    \STATE\COMMENT{Step 1: initialization}
    \STATE $\mathcal{S} \A \emptyset$
    \FOR{$f \A 1$ \TO $F$}
      \STATE $j^* \A \argmax\limits_{j = 1,\dots,M} \eval{\{ k_j^{(f)} \}$}
      \STATE $\mathcal{S} \A \mathcal{S} \cup \{ k_{j^*}^{(f)} \}$
    \ENDFOR

    \REPEAT
      \STATE\COMMENT{Steps 2 and 3: selection of the candidates}
      \STATE $\mathcal{C} \A \emptyset$
      \FOR{$f \A 1$ \TO $F$}
        \STATE $\mathcal{T}^{(f)} \A \{ k_{1}^{(f)}, \dots, k_M^{(f)} \}
                \setminus \mathcal{S}$
        \IF{$\mathcal{T}^{(f)} \neq \emptyset$}
          \STATE $j^* \A \argmax\limits_{j:k_j^{(f)} \in \mathcal{T}^{(f)}}
                  \eval{\mathcal{S} \cup \{ k_j^{(f)} \}}$
          \STATE $\mathcal{C} \A \mathcal{C} \cup \{ k_{j^*}^{(f)} \}$
        \ENDIF
      \ENDFOR

      \STATE\COMMENT{Step 4: selection of the best set of kernels}
      \STATE $\mathcal{B}^* \A \argmax\limits_{\mathcal{B} \subseteq \mathcal{C}}
              \eval{\mathcal{S} \cup \mathcal{B}}$
      \STATE $\mathcal{S} \A \mathcal{S} \cup \mathcal{B}^*$
    \UNTIL{$\mathcal{B}^* = \emptyset$}
  \RETURN $\mathcal{S}$
\end{algorithmic}
\caption{The proposed heuristic: the procedure selects a subset
  $\mathcal{S}$ of the given kernels $\{ k_j^{(f)} \}_{j=1\dots M,
    f=1\dots F}$. Starting from the set of the best kernels
  for each feature, a sequence of iterations selects additional
  kernels.  In each iteration a set $\mathcal{C}$ of candidates is
  formed by taking up to one kernel for each feature vector.  Then the
  best subset $\mathcal{B}^*$ of $\mathcal{C}$ is added to
  $\mathcal{S}$.  The procedure continues until $\mathcal{B}^*$ is
  empty.  The subroutine {\bf evaluate} performs a five-fold cross
  validation on the training set to estimate the goodness of a set of
  kernels.}
  \label{fig:pseudocode}
\end{figure}
The whole procedure is detailed in the pseudo-code reported in
Fig.~\ref{fig:pseudocode}. Step 4 requires the evaluation of up to
$2^F -1$ combinations of candidates, and that step is repeated up to
$M \times F$ times (since at least one kernel is added at each
iteration).  Therefore, in the worst case the number of trained
classifiers is $O(F \times M \times 2^F)$. Such a number can be kept
manageable if the number of features $F$ is reasonable.  As an
example, in the setup used in our experiments $F = 4$ and $M = 9$
therefore the number of classifiers is less than
$4 \times 9 \times 2^4 = 576$, which is several orders of magnitude
less than the $2^{4 \times 9} = 6.87 \times 10^{10}$ combinations
required by the brute force strategy.  As rough indicator consider
that on 10\% of the 19-class dataset described in the next section,
the training with our strategy set required about 45 minutes on a
standard PC.

\section{Experimental evaluation}

To assess the merits of the classifier designed according to our
proposal, we compared it with other approaches in the state of the art
on two data sets. For all the experiments, the ``one versus all'' strategy is used to deal with
multiple classes. The evaluation includes:
\begin{itemize}
\item image features: we evaluated several types of image features and
  their concatenation using SVMs with different kernels (linear, RBF
  and $\chi^2$);
\item MKL: we evaluated three versions of MKL ($p=1$, $p=1.25$, $p=2$
  and the proposed strategy); as alternative heuristic for the
  initialization of MKL we considered the kernel alignement method
  proposed by Tuia~\emph{et al.}~\cite{tuia2010learning} with and without centered kernels~\cite{scholkopf1997kernel};
\item other approaches in the state of the art: we evaluated the
  method proposed by Risojevi\'c \emph{et
    al.}
  ~\cite{risojevic2013fusion} (`metalearner', with both the features
  described in the original paper and the features described in this
  paper).
\end{itemize}

\subsection{Data and image features}

\paragraph*{21-Class Land-Use Dataset} this is a dataset of images of
21 land-use classes selected from aerial orthoimagery with a pixel
resolution of one foot~\cite{yang2010bag}.  For each class, 100 RGB
images at $256 \times 256$ are available.
These classes contain a variety of spatial
patterns, some homogeneous with respect to texture, some homogeneous
with respect to color, others not homogeneous at all.

\paragraph*{19-Class Satellite Scene} this dataset consists of 19
classes of satellite scenes collected from Google Earth (Google
Inc.). Each class has about 50 RGB images, with the size of $600
\times 600$ pixels~\cite{dai2011satellite,xia2010structural}. The
images of this dataset are extracted from very large satellite images
on Google Earth.

We considered four types of image features ($F =
4$): two of them have been taken from the state of the art and have
been chosen because of the good performance that are reported in the
literature.  The other two features have been specially designed,
here, to complement the others.

\subsubsection{Bag of SIFT}
we considered SIFT~\cite{lowe2004distinctive} descriptors computed on
the intensity image and quantized into a codebook of 1096 ``visual
words''.  This codebook has been previously built by $k$-means
clustering the descriptors extracted from more than 30,000 images.  To
avoid unwanted correlations with the images used for the evaluation,
we built the codebook by searching general terms on the flickr web
service and by downloading the returned images.  The feature vector is
a histogram of 1096 visual words.

\subsubsection{Gist}
these are features computed from a wavelet decomposition of
the intensity image~\cite{oliva2001modeling}. Each image location is
represented by the output of filters tuned to different orientations
and scales. This representation is then downsampled to $4
\times 4$ pixels. We used eight orientations and four scales thus, the
dimensionality of the feature vector is $8 \times 4 \times 16 = 512$.

\begin{table*}[tb]
\caption{Performance comparisons on the two datasets.}\label{tab:results}
\begin{center}
\scriptsize
\begin{tabular}{llcccccc}
& &\multicolumn{6}{c}{\bf{\% training images per class (21-class data set)}} \\
\cline{3-8}
\bf{Features}&\bf{Kernel} &5 & 10 & 20 & 50 & 80 & 90 \\
\hline
Bag of SIFT & $\chi^2$&51.89	($\pm$1.73)	&	59.13	($\pm$1.53)	&	66.09	($\pm$1.11)	&	76.80 ($\pm$1.11)	&	77.42	($\pm$1.99)	&	85.45	($\pm$2.83)	\\
Bag of LBP&$\chi^2$&42.07	($\pm$2.35)	&	50.39	($\pm$1.44)	&	59.21	($\pm$1.26)	&	76.51	($\pm$1.16)	&	77.18	($\pm$1.83)	&	74.81	($\pm$2.79)	\\
GIST &$\chi^2$&38.62	($\pm$1.74)	&	45.62	($\pm$1.44)	&	53.73	($\pm$1.20)	&	67.96	($\pm$1.30)	&	68.53	($\pm$1.73)	&	67.35	($\pm$2.63)	\\
LBP of moments&RBF&24.41	($\pm$1.47)	&	29.58	($\pm$1.30)	&	34.54	($\pm$1.05)	&	44.62	($\pm$1.20)	&	45.26	($\pm$2.13)	&	59.49	($\pm$3.24)	\\
\hline
\multirow{2}{*}{Concatenation}&RBF&58.25 ($\pm$2.21)&69.19 ($\pm$1.56) &77.32 ($\pm$1.23)& 85.44 ($\pm$1.04) & 88.91 ($\pm$1.23) & 89.41 ($\pm$1.81)  \\
&$\chi^2$&58.05 ($\pm$2.21)&	68.58	($\pm$1.44)	&	76.97	($\pm$1.21)	&	85.09	($\pm$0.95)	&	88.33	($\pm$1.31)	&	88.55	($\pm$1.92)	\\
\hline
\multirow{4}{*}{Metalearner~\cite{risojevic2013fusion} (original features)} &Ridge regression C&56.05 ($\pm$2.09)& 70.30 ($\pm$1.64)  &79.99 ($\pm$1.21) & 88.45 ($\pm$0.90)& 91.59 ($\pm$1.20)& 91.98 ($\pm$1.74)\\
 &RBF SVM&48.58 ($\pm$2.63)& 64.91 ($\pm$2.13) &79.16 ($\pm$1.23)& \bf{89.28} ($\pm$0.86)& \bf{92.52} ($\pm$1.07)& \bf{92.98} ($\pm$1.64) \\
& RBF ridge regr.&51.77 ($\pm$2.61)& 65.52 ($\pm$2.46) &80.26 ($\pm$1.18)& 88.62 ($\pm$1.01)& 91.76 ($\pm$1.13)& 92.23 ($\pm$1.72)\\
 &RBF ridge regr. C&54.92 ($\pm$2.13)& 69.10 ($\pm$1.94)  &79.79 ($\pm$1.20)& 89.04 ($\pm$0.93)& 92.16 ($\pm$1.11)& 92.57 ($\pm$1.71)\\
 \hline
\multirow{4}{*}{Metalearner~\cite{risojevic2013fusion}} &Ridge regression	& 40.25 ($\pm$2.76)	& 59.88 ($\pm$1.83)	& 71.36 ($\pm$1.32)	& 83.04 ($\pm$1.14)	& 87.79 ($\pm$1.31)	& 88.66 ($\pm$1.99)	\\
&RBF SVM	& 38.75 ($\pm$2.98)	& 52.62 ($\pm$2.39)	& 66.35 ($\pm$1.77)	& 82.72 ($\pm$1.23)	& 88.44 ($\pm$1.43)	& 89.21 ($\pm$1.79)	\\
&RBF ridge regr.	& 42.93 ($\pm$2.71)	& 58.22 ($\pm$2.28)	& 71.78 ($\pm$1.49)	& 83.68 ($\pm$1.11)	& 88.24 ($\pm$1.25)	& 88.81 ($\pm$1.92)	\\
&RBF ridge regr. C	& 39.41 ($\pm$2.37)	& 59.53 ($\pm$1.91)	& 71.52 ($\pm$1.35)	& 83.47 ($\pm$1.06)	& 87.85 ($\pm$1.35)	& 88.58 ($\pm$1.98)	\\
\hline
Kernel Alignment~\cite{tuia2010learning}&Linear, RBF, $\chi^2$& 61.16 ($\pm$1.79)&  	71.42 ($\pm$1.32)&77.65 ($\pm$1.03)& 84.23 ($\pm$0.95)	& 86.88 ($\pm$1.39)& 87.61 ($\pm$2.13)\\
Kernel Alignment~\cite{tuia2010learning} $+$ Kernel Cent.&Linear, RBF, $\chi^2$	&61.59 ($\pm$1.90)	& 70.77 ($\pm$1.70)	& 79.10 ($\pm$1.13)	& 86.56 ($\pm$0.91)	& 90.07 ($\pm$1.29)	& 90.45 ($\pm$1.86)	\\
MKL ($p=1$)~\cite{kloft2011lp}&Linear, RBF, $\chi^2$	& 55.91 ($\pm$2.24)	& 68.28 ($\pm$1.72)	& 78.34 ($\pm$1.16)	& 87.74 ($\pm$0.82)	& 90.50 ($\pm$1.35)	& 91.20 ($\pm$1.74)	\\
MKL ($p=1.25$)~\cite{kloft2011lp}&Linear, RBF, $\chi^2$	& 58.31 ($\pm$2.13)	& 69.42 ($\pm$1.69)	& 78.10 ($\pm$1.20)	& 86.71 ($\pm$0.82)	& 89.70 ($\pm$1.31)	& 90.34 ($\pm$1.76)	\\
MKL ($p=2$)~\cite{kloft2011lp}&Linear, RBF, $\chi^2$	& 59.87 ($\pm$1.96)	& 70.29 ($\pm$1.59)	& 78.12 ($\pm$1.14)	& 86.06 ($\pm$0.84)	& 89.03 ($\pm$1.32)	& 89.69 ($\pm$1.82)	\\
\hline
Proposed MKL + Kernel Normalization &Linear, RBF , $\chi^2$	& 64.07 ($\pm$1.57)	& 74.00 ($\pm$1.50)	& 81.00 ($\pm$1.18)	& 88.50 ($\pm$0.83)	& 91.30 ($\pm$1.22)	& 91.64 ($\pm$1.94)	\\
Proposed MKL&Linear, RBF, $\chi^2$& {\bf 65.20 ($\pm$1.57)} & {\bf 74.57 ($\pm$1.45)} & {\bf 81.11 ($\pm$1.14)} & 88.86 ($\pm$0.90)	& 91.84 ($\pm$1.29) & 92.31 ($\pm$1.78)\\
\hline
\end{tabular}

\vspace{5mm}

\scriptsize
\begin{tabular}{llcccccc}
& &\multicolumn{6}{c}{\bf{\% training images per class (19-class data set)}} \\
\cline{3-8}
\bf{Features}&\bf{Kernel} &5 & 10 & 20 & 50 & 80 & 90 \\
\hline
Bag of SIFT & $\chi^2$&49.94	($\pm$2.91)	&	61.71	($\pm$1.96)	&	71.03	($\pm$1.67)	&	80.32 ($\pm$1.73)	&	84.27	($\pm$2.11)	&	85.45	($\pm$3.26)	\\
LBP&$\chi^2$&39.69	($\pm$2.58)	&	46.24 ($\pm$2.18)	&	56.61	($\pm$1.82)	&	68.88	($\pm$1.79)	&	74.15	($\pm$2.94)	&	74.81	($\pm$3.64)	\\
GIST &$\chi^2$&34.87	 ($\pm$2.71)	&	44.87	($\pm$2.31)	&	53.44	($\pm$1.59)	&	62.03	($\pm$1.73)	&	66.31	($\pm$2.73)	&	67.35	($\pm$3.65)	\\
LBP of moments&RBF&20.43	($\pm$1.85)	&	30.01	($\pm$3.39)	&	46.86	($\pm$2.25)	&	50.12	($\pm$1.89)	&	58.11	($\pm$2.65)	&	59.49	($\pm$4.34)	\\
\hline
\multirow{2}{*}{Concatenation}&RBF&63.28 ($\pm$2.62)&78.28 ($\pm$1.93) &86.66 ($\pm$1.39)& 93.17 ($\pm$1.08) & 95.11 ($\pm$1.47) & 95.90 ($\pm$1.88) \\
&$\chi^2$& 58.03   ($\pm$3.25)& 76.67   ($\pm$2.13)&  86.08   ($\pm$1.61)&  93.26   ($\pm$1.22)& 95.68   ($\pm$1.35)&  96.01   ($\pm$1.99)\\
\hline
\multirow{4}{*}{Metalearner~\cite{risojevic2013fusion} (original features)} &Ridge regression C&49.61 ($\pm$4.24) & 77.03  ($\pm$2.23)  &84.54 ($\pm$1.73)& 92.56 ($\pm$1.01)& 94.93 ($\pm$1.52)& 95.57 ($\pm$1.93)\\
&RBF SVM&41.22 ($\pm$5.06)& 73.14  ($\pm$2.67)  &85.78 ($\pm$2.01)& 95.47 ($\pm$0.88)& \bf{97.22} ($\pm$1.19)& \bf{97.92} ($\pm$1.27) \\
& RBF ridge regr.&44.52	($\pm$5.01)& 76.69  ($\pm$2.42)  &87.24 ($\pm$1.84)& 94.62 ($\pm$1.06)& 96.98 ($\pm$1.33)& 97.60 ($\pm$1.51)\\
&RBF ridge regr. C&44.40    ($\pm$3.95)& 76.13  ($\pm$2.40)  &85.10 ($\pm$1.93)& 93.71 ($\pm$1.20)& 95.69 ($\pm$1.44) & 96.05 ($\pm$1.82)\\
\hline
\multirow{4}{*}{Metalearner~\cite{risojevic2013fusion}} &Ridge regression	& 43.44 ($\pm$4.20)	& 58.89 ($\pm$2.88)	& 77.05 ($\pm$2.08)	& 90.02 ($\pm$1.23)	& 93.20 ($\pm$1.51)	& 93.67 ($\pm$2.34)	\\
&RBF SVM	& 49.74 ($\pm$3.94)	& 69.86 ($\pm$2.72)	& 83.16 ($\pm$1.72)	& 94.80 ($\pm$1.07)	& 96.71 ($\pm$1.12)	& 97.10 ($\pm$1.72)	\\
&RBF ridge regr.	& 53.36 ($\pm$3.65)	& 75.70 ($\pm$2.24)	& 88.06 ($\pm$1.47)	& 95.08 ($\pm$1.00)	& 96.92 ($\pm$1.09)	& 97.01 ($\pm$1.73)	\\
&RBF ridge regr. C	& 34.81 ($\pm$4.28)	& 51.88 ($\pm$3.45)	& 77.70 ($\pm$2.07)	& 90.55 ($\pm$1.29)	& 93.50 ($\pm$1.48)	& 94.02 ($\pm$2.48)	\\
\hline
Kernel Alignment~\cite{tuia2010learning}&Linear, RBF, $\chi^2$&58.27 ($\pm$3.12)&  	74.11 ($\pm$1.84)&82.70 ($\pm$1.55)& 93.65 ($\pm$1.03)	& 94.99 ($\pm$1.40)& 95.98 ($\pm$1.80)\\
Kernel Alignment~\cite{tuia2010learning} $+$ Kernel Cent.&Linear, RBF, $\chi^2$	& 61.50 ($\pm$3.06)	& 78.40 ($\pm$1.94)	& 87.96 ($\pm$1.48)	& 94.62 ($\pm$0.97)	& 96.20 ($\pm$1.15)	& 96.74 ($\pm$1.63)	\\
MKL ($p=1$)~\cite{kloft2011lp}&Linear, RBF, $\chi^2$	& 53.83 ($\pm$3.72)	& 76.11 ($\pm$2.30)	& 87.96 ($\pm$1.42)	& 94.78 ($\pm$0.92)	& 96.50 ($\pm$1.21)	& 96.75 ($\pm$1.67)	\\
MKL ($p=1.25$)~\cite{kloft2011lp}&Linear, RBF, $\chi^2$	& 58.67 ($\pm$3.43)	& 77.34 ($\pm$2.09)	& 87.61 ($\pm$1.46)	& 94.53 ($\pm$0.89)	& 96.31 ($\pm$1.12)	& 96.70 ($\pm$1.73)	\\
MKL ($p=2$)~\cite{kloft2011lp}&Linear, RBF, $\chi^2$	& 61.88 ($\pm$3.01)	& 78.49 ($\pm$2.00)	& 87.64 ($\pm$1.48)	& 94.21 ($\pm$0.93)	& 95.95 ($\pm$1.19)	& 96.53 ($\pm$1.86)	\\
\hline
Proposed MKL + Kernel Normalization &Linear, RBF, $\chi^2$	& 69.73 ($\pm$2.67)	& 83.45 ($\pm$1.63)	& 90.39 ($\pm$1.23)	& 95.33 ($\pm$0.78)	& 96.60 ($\pm$1.01)	& 96.90 ($\pm$1.53)	\\
Proposed MKL&Linear, RBF, $\chi^2$& {\bf 70.20	($\pm$2.54)}	& {\bf 84.05	($\pm$1.67)}	& {\bf 90.80	($\pm$1.26)}	& {\bf 95.73	($\pm$0.85)}	& 96.83	($\pm$1.04)	& 97.36	($\pm$1.48)	\\
\hline
\end{tabular}
\end{center}
\end{table*}

\subsubsection{Bag of dense LBP}
local binary patterns are calculated on square patches of size $w
\times w$ that are extracted as a dense grid from the original
image. The final descriptor is obtained as a bag of such LBP patches
obtained from a previously calculated dictionary. As for the SIFT, the
codebook has been calculated from a set of thousands of generic scene
images.  Differently from SIFT, here we used a grid sampling with a
step of 16 pixels.  Note that this descriptor has been computed
separately on the RGB channels and then concatenated.  We have chosen
the LBP with a circular neighbourhood of radius 2 and 16 elements, and
18 uniform and rotation invariant patterns. We set $w = 16$ and $w =
30$ for the 21-classes and 19-classes respectively. The size of the
codebook, as well as the size of the final feature vector, is 1024.

\subsubsection{LBP of dense moments}
the original image is divided in a dense grid of
$N$ square patches of size $w \times w$. The mean and the standard
deviation of each patch is computed for the red, green and blue
channels.  Finally, the method computes the LBP of each matrix and the
final descriptor is then obtained by concatenating the two resulting
LBP histograms of each channel. We have chosen the LBP with a circular
neighbourhood of radius 2 and 16 elements, and 18 uniform
and rotation invariant patterns. We set $w = 16$ and $w = 30$ for the
21-classes and 19-classes respectively. The final dimensionality of
the feature vector is $3 \times (18 + 18) = 108$.

\subsection{Experiments}

For each dataset we used training sets of different sizes.  More in
detail, we used 5, 10, 20, 50, 80 and 90\% of the images for training the methods, and the
rest for their evaluation.  To make the results as robust as possible,
we repeated the experiments 100 times with different random partitions
of the data (available on the
authors' web
page\footnote{\url{http://www.ivl.disco.unimib.it/research/hmkl/}}).

In the experiments for single kernel we used the parallel version of
LIBSVM\footnote{\url{http://www.maths.lth.se/matematiklth/personal/sminchis/code/}}
~\cite{li2010holistic}. For MKL we used the algorithm proposed by
Sonnenburg \emph{et al.}~\cite{kloft2009efficient}\cite{kloft2011lp}
that is part of the SHOGUN
toolbox. 
For both single and multiple kernel experiments, we considered the
linear, Gaussian RBF and $\chi^2$ kernel functions.  In the case of
single kernel the parameter $\gamma$ has been found by the standard
SVM model selection procedure; for MKL we used the values
$\gamma = \{ 10,1,0.1,0.01\}$ and $\gamma = \{ 3,2,1,0.5\}$ for the
Gaussian RBF and $\chi^2$ kernels respectively (chosen taking into
account the average Euclidean and $\chi^2$ distances of the features;
more values could be used at the expenses of an increase in
computation time).  The coefficient $C$ has been chosen among the
values $\{ 0.1, 1, 2, 3, 4, 5 \}$. For experiments with the
metalearner we used the code kindly provided by the authors without
any modifications. For all the other methods, before the computation
of the kernels, all the feature vectors, have been $L_2$
normalized. For the standard MKL and the heuristic MKL we also
normalized all the kernels by the standard deviation in the Hilbert's
space.

Table~\ref{tab:results} reports the results obtained.
Among single features, bag of SIFT obtained the
highest accuracy, with the exception of the case of a large training
set for the 21-class dataset where the best feature is the bag of LBP
(for the sake of brevity, only the results obtained with the best
kernel are reported).  Regardless the size of the training set, the
concatenation of the feature vectors improved the results.  In fact, features such as the LBP of moments
that resulted weak when used alone, demonstrated to be useful when
combined with other features.

On the 21-class dataset advanced combination strategies, such as the
metalearners proposed in~\cite{risojevic2013fusion} or the MKL as
defined in~\cite{kloft2011lp}, performed better than simple
concatenations.  However, this is not true for the 19-class
dataset. The metalearner works better with its own original features
than with our features (this could be due to the fact that these
features were based on a codebook defined on the same dataset).

The proposed MKL strategy, in the case of small training sets,
outperformed all the other methods considered.  More in detail, when
trained on 10\% of the 21-class dataset the accuracy of our method was
by at least 3\% better than the accuracy of the other strategies.
Similarly, for the 10\% of the 19-class dataset, our method was the
only one with an accuracy higher than 80\%.  Even larger improvements have been obtained when using 5\% of the data for training. The improvement obtained
in the case of small training sets is statistically significant as it
passed a two-tailed paired t-test with a confidence level of 95\%.

\begin{figure}[t]
  \centering
  \setlength{\tabcolsep}{1pt}
  \includegraphics[width=0.35\textwidth]{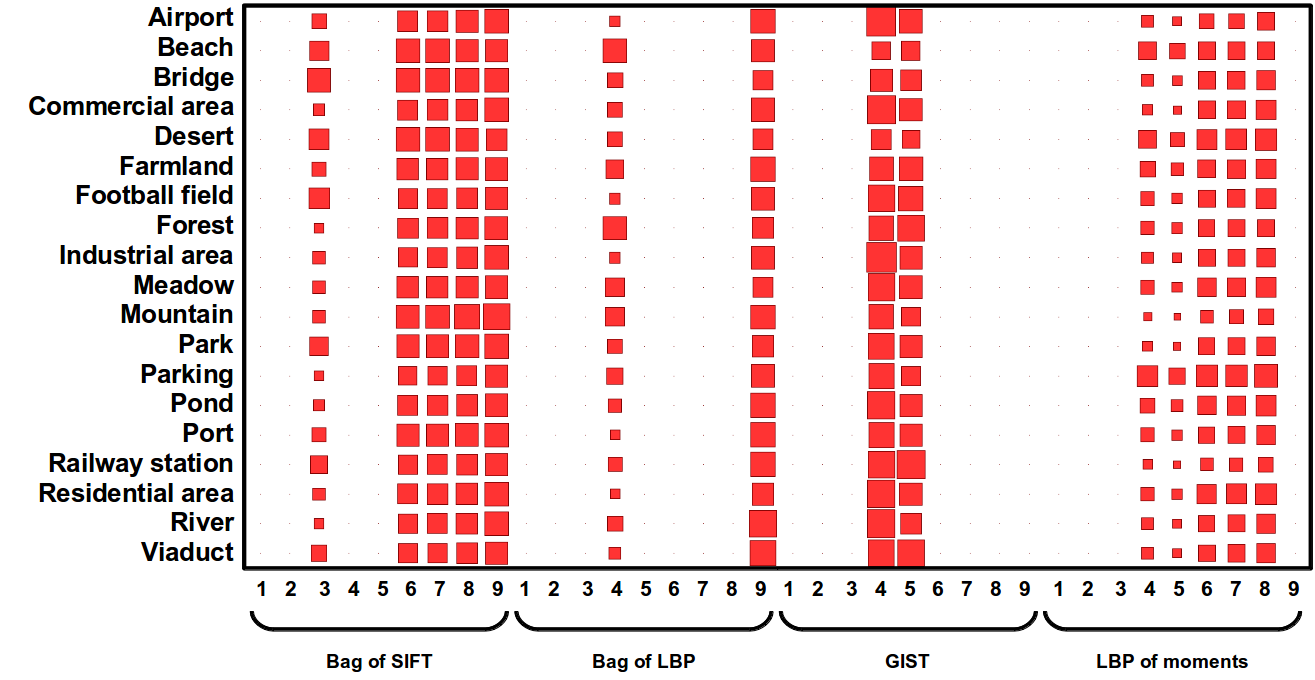}\\
  \caption{Weights selected by the heuristic
    MKL when trained on the 19-class data set.
    Numbers at the bottom indicate kernels: 1 for linear, 2--5
    for RBFs at different $\gamma$, 6--9 for $\chi^2$s at different
    $\gamma$. The size of the squares is proportional to the weights.
    In order to make the weights comparable, we preprocessed
    the kernels with the normalization scheme proposed by Zien and
    Ong~\cite{zien2007multiclass}.
  }
  \label{fig:mkl_weights}
\end{figure}

\begin{figure}[t]
  \centering
  \setlength{\tabcolsep}{1pt}
  \def\arraystretch{0.4}%
  \includegraphics[width=0.26\textwidth]{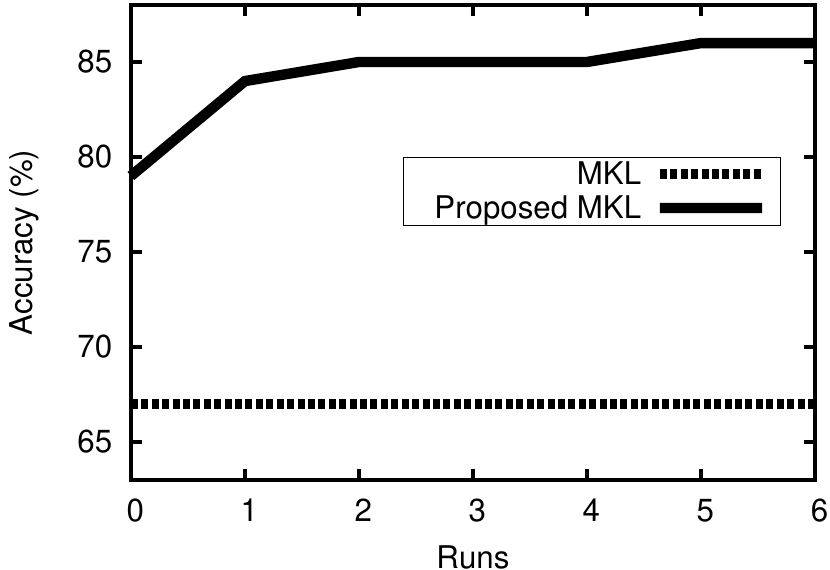}\\
  \caption{Accuracy of the proposed MKL varying the number of
    iterations, on the 19-class data set with 10\% of training images.  The performance of standard MKL
    ($p=2$) is reported for comparison.}
  \label{fig:mkl_iterations}
\end{figure}

The weights assigned by the heuristic MKL are quite stable across the
different classes (see Figure~\ref{fig:mkl_weights}).
Figure~\ref{fig:mkl_iterations} shows the classification accuracy as a
function of the number of iterations of the proposed method when 10\%
of images are used for training.  According to the plot most of the
improvement with respect to standard MKL is obtained after just a
couple of iterations.

\section{Conclusions}

In land use classification scenes may appear at different scales, orientations and lighting conditions. Therefore, multiple features must be combined in order to obtain good performance.
In this paper we presented a classification strategy that improves MKL by making it suitable for small training
sets.  Experimental results on two public land use datasets shown that our method performs better than the other alternatives considered.
We believe that our approach could be applied to other image classification
tasks.  In particular, we expect that it may be successful in those
cases where few training data are available and, at the same time,
multiple features are needed.

Our MKL strategy may be applied to any set of features, future works
will address a deeper investigation about feature design. It has been
demonstrated that kernel normalization techniques can significantly
influence the performance of MKL~\cite{kloft2011lp}, we will also
investigate how to effectively exploit these techniques within our MKL
strategy.

\section*{Acknowledgment}
We would like to thank Vladimir Risojevi{\'c} and Zdenka Babi{\'c} for making available the code of their method.

\ifCLASSOPTIONcaptionsoff
  \newpage
\fi

\bibliography{GRSL-2014}

\begin{thebibliography}{10}
\providecommand{\url}[1]{#1}
\csname url@samestyle\endcsname
\providecommand{\newblock}{\relax}
\providecommand{\bibinfo}[2]{#2}
\providecommand{\BIBentrySTDinterwordspacing}{\spaceskip=0pt\relax}
\providecommand{\BIBentryALTinterwordstretchfactor}{4}
\providecommand{\BIBentryALTinterwordspacing}{\spaceskip=\fontdimen2\font plus
\BIBentryALTinterwordstretchfactor\fontdimen3\font minus
  \fontdimen4\font\relax}
\providecommand{\BIBforeignlanguage}[2]{{%
\expandafter\ifx\csname l@#1\endcsname\relax
\typeout{** WARNING: IEEEtran.bst: No hyphenation pattern has been}%
\typeout{** loaded for the language `#1'. Using the pattern for}%
\typeout{** the default language instead.}%
\else
\language=\csname l@#1\endcsname
\fi
#2}}
\providecommand{\BIBdecl}{\relax}
\BIBdecl

\bibitem{campbell2002introduction}
J.~Campbell, \emph{Introduction to remote sensing}.\hskip 1em plus 0.5em minus
  0.4em\relax CRC Press, 2002.

\bibitem{sheng2012highresolution}
G.~Sheng, W.~Yang, T.~Xu, and H.~Sun, ``High-resolution satellite scene
  classification using a sparse coding based multiple feature combination,''
  \emph{Int'l J. of Remote Sensing}, vol.~33, no.~8, pp. 2395--2412, 2012.

\bibitem{risojevic2011gabor}
V.~Risojevi{\'c}, S.~Momi{\'c}, and Z.~Babi{\'c}, ``Gabor descriptors for
  aerial image classification,'' in \emph{Adap. \& Nat. Comp. Alg.}, 2011, pp.
  51--60.

\bibitem{risojevic2013fusion}
V.~Risojevic and Z.~Babic, ``Fusion of global and local descriptors for remote
  sensing image classification,'' \emph{Geoscience and Remote Sensing Letters},
  vol.~10, no.~4, pp. 836--840, 2013.

\bibitem{shao2013hierarchical}
W.~Shao, W.~Yang, G.-S. Xia, and G.~Liu, ``A hierarchical scheme of multiple
  feature fusion for high-resolution satellite scene categorization,'' in
  \emph{Computer Vision Systems}, 2013, pp. 324--333.

\bibitem{yang2010bag}
Y.~Yang and S.~Newsam, ``Bag-of-visual-words and spatial extensions for
  land-use classification,'' in \emph{Proc. of the Int'l Conf. on Advances in
  Geographic Information Systems}, 2010, pp. 270--279.

\bibitem{dai2011satellite}
D.~Dai and W.~Yang, ``Satellite image classification via two-layer sparse
  coding with biased image representation,'' \emph{Geoscience and Remote
  Sensing Letters}, vol.~8, no.~1, pp. 173--176, 2011.

\bibitem{lanckriet2004statistical}
G.~R. Lanckriet, T.~De~Bie, N.~Cristianini, M.~I. Jordan, and W.~S. Noble, ``A
  statistical framework for genomic data fusion,'' \emph{Bioinformatics},
  vol.~20, no.~16, pp. 2626--2635, 2004.

\bibitem{sonnenburg2006large}
S.~Sonnenburg, G.~R{\"a}tsch, C.~Sch{\"a}fer, and B.~Sch{\"o}lkopf, ``Large
  scale multiple kernel learning,'' \emph{J. Mach. Learn. Res.}, vol.~7, pp.
  1531--1565, 2006.

\bibitem{tuia2010learning}
D.~Tuia, G.~Camps-Valls, G.~Matasci, and M.~Kanevski, ``Learning relevant image
  features with multiple-kernel classification,'' \emph{IEEE Trans. on Geosci.
  Remote Sens.}, vol.~48, no.~10, pp. 3780--3791, 2010.

\bibitem{subrahmanya2010sparse}
N.~Subrahmanya and Y.~C. Shin, ``Sparse multiple kernel learning for signal
  processing applications,'' \emph{IEEE Trans. Pattern Anal. Mach. Intell.},
  vol.~32, no.~5, pp. 788--798, 2010.

\bibitem{gu2012representative}
Y.~Gu, C.~Wang, D.~You, Y.~Zhang, S.~Wang, and Y.~Zhang, ``Representative
  multiple kernel learning for classification in hyperspectral imagery,''
  \emph{IEEE Trans. on Geosci. Remote Sens.}, vol.~50, pp. 2852--2865, 2012.

\bibitem{kloft2011lp}
M.~Kloft, U.~Brefeld, S.~Sonnenburg, and A.~Zien, ``Lp-norm multiple kernel
  learning,'' \emph{J. Mach. Learn. Res.}, vol.~12, pp. 953--997, 2011.

\bibitem{Lanckriet2004}
G.~R.~G. Lanckriet, N.~Cristianini, P.~Bartlett, L.~E. Ghaoui, and M.~I.
  Jordan, ``Learning the kernel matrix with semidefinite programming,''
  \emph{J. Mach. Learn. Research}, vol.~5, pp. 27--72, 2004.

\bibitem{kloft2009efficient}
M.~Kloft, U.~Brefeld, S.~Sonnenburg, P.~Laskov, K.-R. M{\"u}ller, and A.~Zien,
  ``Efficient and accurate lp-norm multiple kernel learning,'' in
  \emph{Advances in NIPS}, vol.~22, no.~22, 2009, pp. 997--1005.

\bibitem{scholkopf1997kernel}
B.~Sch{\"o}lkopf, A.~Smola, and K.-R. M{\"u}ller, ``Kernel principal component
  analysis,'' in \emph{ICANN'97}.\hskip 1em plus 0.5em minus 0.4em\relax
  Springer, 1997, pp. 583--588.

\bibitem{xia2010structural}
G.-S. Xia, W.~Yang, J.~Delon, Y.~Gousseau, H.~Sun, H.~Ma{\^\i}tre
  \emph{et~al.}, ``Structural high-resolution satellite image indexing,'' in
  \emph{ISPRS TC VII Symposium-100 Years ISPRS}, vol.~38, 2010, pp. 298--303.

\bibitem{lowe2004distinctive}
D.~Lowe, ``Distinctive image features from scale-invariant keypoints,''
  \emph{Int'l J. Computer Vision}, vol.~60, no.~2, pp. 91--110, 2004.

\bibitem{oliva2001modeling}
A.~Oliva and A.~Torralba, ``Modeling the shape of the scene: A holistic
  representation of the spatial envelope,'' \emph{Int'l J. Computer Vision},
  vol.~42, no.~3, pp. 145--175, 2001.

\bibitem{li2010holistic}
F.~Li, J.~Carreira, and C.~Sminchisescu, ``Object recognition as ranking
  holistic figure-ground hypotheses,'' in \emph{IEEE CVPR}, 2010.

\bibitem{zien2007multiclass}
A.~Zien and C.~S. Ong, ``Multiclass multiple kernel learning,'' in \emph{Proc.
  of the Int'l Conf. on Mach. learning}, 2007, pp. 1191--1198.

\end{thebibliography}
\bibliographystyle{IEEEtran}

%

%
%
%




\end{document}